\documentclass[10pt,twocolumn,letterpaper]{article}

\usepackage{iccv}
\usepackage{times}
\usepackage{epsfig}
\usepackage{graphicx}
\usepackage{amsmath}
\usepackage{amssymb}
\usepackage{bbm}
\usepackage{paralist} %
\usepackage{caption}  %
\usepackage{subcaption}  %
\usepackage{xcolor}  %
\usepackage{multirow}  %
\usepackage{booktabs}  %
\usepackage[numbers,sort&compress]{natbib}  %
\usepackage{arydshln}

\usepackage{subcaption}
\usepackage{dutchcal}
\usepackage[pagebackref=true,breaklinks=true,letterpaper=true,colorlinks,bookmarks=false]{hyperref}

\iccvfinalcopy %

\def\iccvPaperID{3422} %

\ificcvfinal\fi

\begin{document}
\title{Adversarial Motion Modelling helps Semi-supervised Hand Pose Estimation}

\author{Adrian Spurr$^{1}$ \quad Pavlo Molchanov$^{2}$ \quad Umar Iqbal$^{2}$ \quad Jan Kautz$^{2}$ \quad Otmar Hilliges$^{1}$\vspace{0.1cm} \\
 $^1$ETH Zurich \quad
 $^2$NVIDIA \\
}

\maketitle

\newcommand{\AS}[1]{{\color{red}[AS: #1]}}
\newcommand{\PM}[1]{{\color{orange}[PM: #1]}}
\newcommand{\UI}[1]{{\color{pink}[UI: #1]}}
\newcommand{\OH}[1]{{\color{blue}[OH: #1]}}
\newcommand{\JK}[1]{{\color{cyan}[JK: #1]}}

\newcommand{\otmar}[1]{\OH{#1}}
\newcommand{\oh}[1]{\OH{#1}}
\newcommand{\pmnote}[1]{\PM{#1}}

\newcommand{\todo}[1]{{\color{cyan}\textbf{TODO}: #1}}
\newcommand{\newtext}[1]{{\color{magenta}\textbf{NEW}: #1}}
\newcommand{\head}[1]{\noindent\textbf{#1}}

\newcommand{\figref}[1]{Fig.~\ref{#1}}
\newcommand{\tabref}[1]{Tab.~\ref{#1}}
\newcommand{\V}[1]{\mathbf{#1}}
\newcommand{\R}[0]{\rm I\!R}
\newcommand{\E}[0]{\rm I\!E}
\newcommand{\loss}[0]{\mathcal{L}}

\newcommand{\Rone}{\textbf{\color{Thistle}{R1}}}
\newcommand{\Rtwo}{\textbf{\textcolor{ForestGreen}{R2}}}
\newcommand{\Rthree}{\textbf{\textcolor{NavyBlue}{R3}}}
\newcommand{\A}{\textbf{A:}}
\begin{abstract}
Hand pose estimation is difficult due to different environmental conditions, object- and self-occlusion as well as diversity in hand shape and appearance. Exhaustively covering this wide range of factors in fully annotated datasets has remained impractical, posing significant challenges for generalization of supervised methods. 
Embracing this challenge, we propose to combine ideas from adversarial training and motion modelling to tap into unlabeled videos. To this end we propose what to the best of our knowledge is the first motion model for hands and show that an adversarial formulation leads to better generalization properties of the hand pose estimator via semi-supervised training on unlabeled video sequences. 
In this setting, the pose predictor must produce a valid sequence of hand poses, as determined by a discriminative adversary. 
This adversary reasons both on the structural as well as temporal domain, effectively exploiting the spatio-temporal structure in the task.
The main advantage of our approach is that we can make use of unpaired videos and joint sequence data both of which are much easier to attain than paired training data.
We perform extensive evaluation, investigating essential components needed for the proposed framework and empirically demonstrate in two challenging settings that the proposed approach leads to significant improvements in pose estimation accuracy. In the lowest label setting, we attain an improvement of $40\%$ in absolute mean joint error.

\end{abstract}
\section{Introduction}
\label{sec:intro}
\begin{figure}[t]
    \centering
    \includegraphics[width=\columnwidth]{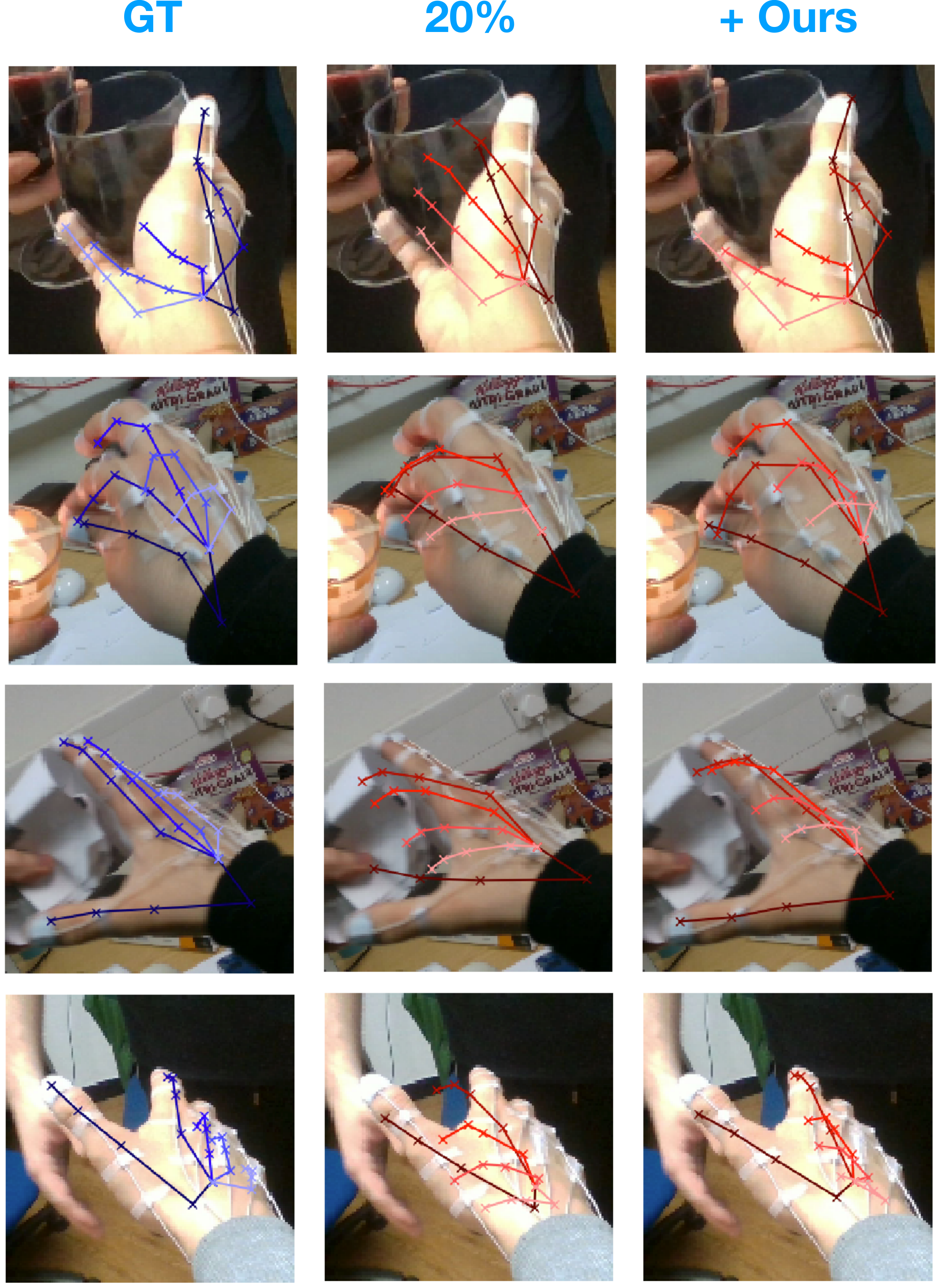}
    \caption{\textbf{Adversarial motion modelling improves hand pose estimation} via leveraging \emph{unlabeled} videos. (Left) Ground-truth labels. (Middle) Given only a fraction of labeled data, standard hand pose models perform poorly on unseen samples. (Right) Using our proposed adversarial motion model, we train the per-frame hand pose model to produce predictions on unlabeled videos of valid motion, leading to significantly more accurate pose estimation.} %
    \label{fig:qual_results_teaser}
\end{figure}

Estimating the 3D hand pose from monocular RGB is in itself a challenging task. Many application areas such as VR/AR require that hand pose estimation models are robust to different environmental conditions, object- and self-occlusion as well as diversity in hand shape and texture. Although impressive performance has been achieved by supervised learning approaches (e.g. \cite{boukhayma20193d, baek2020weakly, hasson2020leveraging}), this has only been demonstrated in constrained environments. This limitation stems from the need for annotated data, which is typically only available if conditions can be controlled due to the difficulties to acquire 3D hand pose ground truth. However, as \cite{zimmermann2019freihand} demonstrated, models trained on labeled datasets typically do not generalize well to other settings. 

One option would be to acquire more labeled data in a more diverse range of settings. Yet, such attempts would need to overcome many challenges since correctly annotating complex and diverse 3D hand poses is difficult even with sophisticated ground truth acquisition methods. In addition, such setups limit portability and hence the setting. This can be partially mitigated by making use of weakly supervised data \cite{cai2018weakly, boukhayma20193d, spurr2020weakly} or machine-annotations. For example, \cite{kulon2020weakly} introduce a new dataset by making use of a machine-annotation method that samples videos from YouTube containing hands. These sequences are labeled using OpenPose \cite{cao2019openpose} and by fitting the MANO \cite{romero2017mano} model to the obtained 2D keypoints. While the idea to leverage vast amounts of YouTube videos is promising, the labelling approach is inherently bounded by the accuracy of OpenPose. 

We ask the question, if there is an alternative approach to make use of unlabeled videos. For this, we take inspiration from the field of learned motion modeling. Research in that area focuses on modeling the 3D motion of human bodies, by either performing motion-infilling \cite{kaufmann2020convolutional, holden2015learning, harvey2020robust, hernandez2019human}, extrapolation \cite{aksan2019structured, hernandez2019human,martinez2017human,cai2020learning,Wang2018Adversarial,jain2016structural,Pavllo2018BMVC,mao2019learning,mao2020history} or denoising \cite{holden2015learning, ghosh2017learning} by exploiting temporal information linking the individual poses of a given sequence. 

The predictions of a hand pose estimator on videos can be interpreted as a sequence of hand motion.
Thus it is feasible to make use of learned motion modeling in order to exploit temporal information and to correct per-frame predictions by enforcing consistency and validity of the motion of the predicted hand poses. However, we observe that so far there is no prior work on motion modeling of hand poses. 
Unpacking this observation further we note that full body and hand motion have important differences. Full body motion is often cyclic and body motion models tend to exploit this \cite{aksan2020attention, mao2020history}. Additionally, the motion modeling efforts for body pose are driven by the availability of datasets of appropriate size, such as Human3.6M \cite{ionescu2013human3} and AMASS \cite{mahmood2019amass}. In contrast, hand motion is less cyclic and can contain changes in pose over a very short time horizon. Potentially because of these reasons, simply applying existing motion-models to hand pose estimates is not straightforward and in our experiments did not yield sufficiently good performance. Additionally, there is a lack of suitable hand motion datasets.

Sequential hand pose datasets such as BigHands2.2M \cite{yuan2017bighand2} do exist. However, participants where instructed to explore the full range of motion of hands, including extremal poses, and to perform random movement. Hence its motion statistics differ significantly from natural motion and the distribution of joint velocity/acceleration is significantly different from other datasets in the literature, see supplementary for details. In our experiments the inclusion of BH2.2M has lead to detrimental results.

We therefore propose a simple, yet effective way to leverage temporal information from unlabeled videos via an adversarial motion model. Adversarial training has shown promising results in the field of body pose estimation \cite{hmrKanazawa17, kanazawa2019learning, kocabas2019vibe}, but has not yet been explored on the hand pose estimation task. %

The goal of the adversarial setting is for the pose predictor to learn to produce a valid sequence of hand poses on unlabeled videos, as determined by the discriminator. Analogously, the discriminator is trained to distinguish between ground-truth sequences and predictions. We empirically demonstrate that optimizing this min/max game leads to significant performance gains for the hand pose model accuracy. Importantly, at inference time we do not require sequence data and hence the adversarial motion model can be used to improve existing per-frame hand pose models. 

The main advantage of making use of motion models for learning from unlabeled videos is its unpaired nature. Hence, our model does not require fully labeled video sequences. Instead we only require sequences of hand motion to be available as well as sequences of videos containing hands. Such a setting not only has its uses in semi-supervised learning, but could also be used in scenarios where video and motion are both recorded, but synchronization or calibration are infeasible.

In this paper, we demonstrate some essential components in making a hand pose model learn from unlabeled videos using a discriminator. For each component, we provide empirical evidence to support its use. In the interest of future research, we also report building blocks that lead to detrimental results. We envision that our work will provide the basis for follow-ups on applying an adversarial approach to hand pose estimation.
We evaluate our model in two challenging semi-supervised settings on the FPHAB \cite{garcia2018first} and HO-3D \cite{hampali2019ho} dataset and demonstrate how adversarial learning improves the performance of the hand pose model. In lowest-label settings, we observe improvements of up to $40\%$ in absolute mean joint error. %

In summary, our contribution are as follow. \begin{inparaenum}[i)]
\item We introduce a simple and practical method to motion modeling for hands using an adversarial approach
\item We provide empirical evidence for the design choices of said motion model, providing a solid foundation on which future work can build on.
\item We make use of this acquired knowledge to tap into the area of semi-supervised learning from unlabeled video data and 
\item show that our approach leads to significant improvements for the hand pose model.
\end{inparaenum}

\section{Related work}
Our method aims to learn from unlabeled videos, making use of adversarial models to model motion. Here we briefly review the literature on learning-based hand pose estimation approaches. We then continue focusing on adversarial methods in context of pose estimation. Lastly, we briefly discuss the area of motion models to which the discriminator used in our approach pertains.

\noindent\textbf{Hand pose estimation.}
Several approaches have been introduced to perform learning-based hand pose estimation. These generally predict 3D joint skeletons directly \cite{zb2017hand, spurr2018cross, mueller2018ganerated, iqbal2018hand, cai2018weakly, tekin2019ho, yang2019disentangling, doosti2020hope, spurr2020weakly, moon2020interhand2}, follow the MANO paradigm \cite{romero2017mano}, where the parameters of a parametric hand model are regressed \cite{baek2019pushing, boukhayma20193d, hasson2019learning, baek2020weakly, hasson2020leveraging, zhang2019end}, or predict the full mesh model of the hand directly \cite{ge20193d, kulon2020weakly, moon2020deephandmesh}. In the following, we will detail these works. Predicting 3D joint skeleton directly tends to achieve higher accuracy, however they do not provide dense surfaces. \cite{zb2017hand} introduce a staged approach where the 2D keypoints are regressed and then lifted to 3D. \cite{mueller2018ganerated} create a synthetic dataset and reduce the synthetic/real discrepancy via a GAN. \cite{spurr2018cross} propose a cross-modal latent space to facilitate better learning. \cite{iqbal2018hand} presents a 2.5D hand representation, achieving state-of-the-art results. In our work, we use the same representation. \cite{cai2018weakly} augment the training by making use of supplementary depth supervision. \cite{tekin2019ho} present a unified approach for hand and object pose estimation as well as action recognition. \cite{yang2019disentangling} introduce a disentangled latent space to perform better image synthesis. \cite{doosti2020hope} use a graph-based refinement network to jointly refine the object and hand pose predictions. \cite{spurr2020weakly} introduce a biomechanical model to better refine the pose predictions on weakly-supervised data. \cite{moon2020interhand2} address the issue of inter-hand interaction by proposing a model that can predict the pose of both hands at the same time. 

Template-based approaches like MANO implicitly induce a prior of poses upon the predictive model and provide a mesh surface. However due to the regularization of its parameters, its representation space is limited. \cite{baek2019pushing, boukhayma20193d, zhang2019end} predict the MANO parameters directly, making use of additional weak supervision such as hand masks \cite{baek2019pushing, zhang2019end} or in-the-wild 2D annotations \cite{boukhayma20193d, zhang2019end}. \cite{hasson2019learning} jointly learns the MANO mesh, as well as the object mesh for hand-object pose estimation. \cite{hasson2020leveraging} extends this framework in order to learn from partially labeled sequences exploiting a photometric loss on the unlabeled frames. However, they assume the object mesh \emph{to be known a priori} and only estimate its 6D pose. In terms of labeling setting, it is the closest to ours. \cite{moon2020deephandmesh} propose an alternative to MANO by predicting pose and subject dependant correctives to base hand model. 

Regressing the dense surface of a hand directly is the most generalizable approach to obtaining a mesh, however it requires  corresponding annotations which may be difficult to acquire. \cite{ge20193d} alleviate this by introducing a synthetic dataset that is fully annotated with the corresponding mesh and perform noisy supervision for real data. \cite{kulon2020weakly} regress the mesh directly using spiral-convolution and supervise their approach using a MANO model.

\noindent\textbf{Generative Adversarial Nets.}
Generative adversarial networks \cite{goodfellow2014generative} have been used in the body pose literature to refine pose predictions. \cite{hmrKanazawa17, kanazawa2019learning} use a discriminator to distinguish if the predicted SMPL parameters correspond to a real human pose. \cite{kocabas2019vibe} extend the adversarial model to distinguish between valid and predicted sequences of SMPL parameters. This method is the most similar to ours in that they make use of adversarial loss that operates on sequences of poses. We differ from them in two main aspects. First, our task is hand pose estimation which contains different and irregular sequence of motions. Second, our setting considers learning from unlabeled videos whereas \cite{kocabas2019vibe} aim to refine predictions in the fully supervised setting.

\noindent\textbf{Motion modeling.} Our proposed adversarial approach can be categorized as a motion model. Modeling human motion has been in the focus of many approaches in the literature. To the best of our knowledge, no such motion model exists for hand motion. Most body pose motion models follow either a recurrent modeling approach \cite{aksan2019structured, fragkiadaki2015recurrent, ghosh2017learning, jain2016structural, martinez2017human, pavllo2018quaternet, gui2018adversarial, zhou2018auto} or non-recurrent approach, instead opting to make use of graph convolutional networks or convolution networks \cite{kaufmann2020convolutional, li2018convolutional, hernandez2019human}. As our discriminative method runs on a convolutional architecture, we will focus on such works here. \cite{li2018convolutional, kaufmann2020convolutional} use an encoder-decoder layer to predict motion. \cite{hernandez2019human} introduce special encoding-decoding layers to alleviate the lack of spatial continuity of the matrix representation of human pose sequences. Some motion models also employ an adversarial loss \cite{hernandez2019human, gui2018adversarial, barsoum2018hp} to regularize the pose predictions. Similarly, we also use an adversarial model to learn on unlabeled videos and show that a simple approach yields promising results.
\section{Method}

\begin{figure*}[t]
    \centering
    \includegraphics[width=0.99\textwidth]{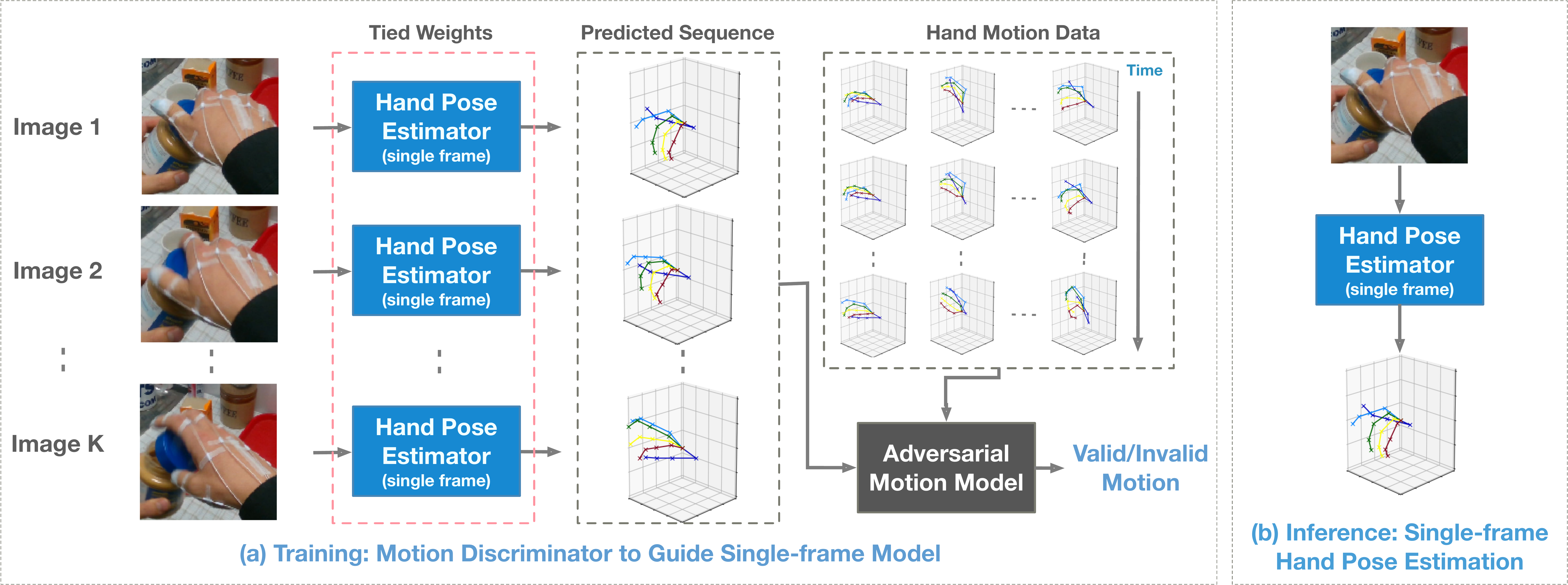}
    \caption{\textbf{Method overview.} a) Given the frames of an unlabeled video sequence, our hand pose model predicts the joints of a hand  skeleton. The per-frame predictions are concatenated into a sequence, generating a hand motion. The predicted motion sequence is input into an adversarial motion model which is capable of discriminating between plausible and invalid hand motion. This capability is learned via unpaired hand motion data. The hand pose estimator is then trained to produce valid motions using the gradients of the discriminator. b) Our method is general in the sense that during inference time, the hand pose estimator only requires a single frame to predict the corresponding pose.}
    \label{fig:method_overview}
\end{figure*}

We introduce our method to leverage unlabeled videos via an adversarial motion model to improve the performance of a hand pose estimation model. Our approach is summarized in \figref{fig:method_overview}. Given a pre-trained hand pose estimation model, we predict for each frame of a video sequence a hand pose. Predictions are concatenated to create a motion sequence. This is input into a motion model that is capable of determining the validity of the given sequence. This capability is learned with the help of unpaired ground-truth hand motion data. The gradient feedback of the motion model helps the hand pose model in improving its prediction performance on the unlabeled videos. Our key contribution is the introduction of the necessary building blocks in order to achieve learning on unlabeled videos.

\noindent\textbf{Notation.} We denote variables $x$ that are outputs of a network via the hat notation $\hat{x}$, lowercase boldface denote vectors $\V{x}$, whereas uppercase boldface denote matrices $\V{X}$.

\subsection{Hand pose model}
Our hand pose model $M$ uses the 2.5D representation proposed by \cite{iqbal2018hand}, where the network predicts both the 2D keypoints $\mathbf{J}^{2D} \in \R^{21\times2}$ and the root-relative depth $\mathbf{z}^r \in \R^{21}$ of a monocular RGB image. The relationship between $\mathbf{J}^{3D}$ and $\mathbf{J}^{2D}, \mathbf{z}^r$ is expressed as follows:
\begin{equation}
    \begin{split}
    \label{eq:3d25d}
    \V{J}^{2D} &= \frac{1}{\V{z}} \V{K} \V{J}^{3D} \\
    \V{z}^r &= \V{z} -  z^{\mathrm{root}}
    \end{split}
\end{equation}
Where $\V{K}$ is the camera intrinsics and $z^{\mathrm{root}}$ is the absolute depth value of the root joint (i.e $\V{j}^{3D}_{\mathrm{root}} = [x^{\mathrm{root}}, y^{\mathrm{root}}, z^{\mathrm{root}}]$). Both branches are trained on the available labeled data using the L1 norm:
\begin{equation}
    \begin{split}
    \label{eq:loss_25d}
    \loss_{J^{2D}} &= |\mathbf{\hat{J}}^{2D} - \mathbf{J}^{2D}| \\
    \loss_{z^r} &= |\mathbf{\hat{z}}^r - \mathbf{z}^r| 
    \end{split}
\end{equation}
As detailed in \cite{iqbal2018hand}, both values $\mathbf{\hat{J}}^{2D}$ and $\mathbf{\hat{z}}^r$ can be used to acquire the 3D pose $\mathbf{J}^{3D} \in \R^{21\times3}$. Prior work \cite{spurr2020weakly} emphasized that this acquisition step can be unstable and introduce a refinement step to alleviate this which we also use here. The resulting 3D pose is recovered as follows:
\begin{equation}
    \begin{split}
    \label{eq:25d3d}
    \V{J}^{3D} = \V{K}^{-1} \V{J}^{2D} (\V{z}^r + z^{\mathrm{root}}_{\mathrm{ref}})
    \end{split}
\end{equation}
Here $z^{\mathrm{root}}_{\mathrm{ref}}$ is the \textit{refined} absolute depth value. 

Before performing adversarial learning on the unlabeled video, we pre-train $M$ on the labeled data. We emphasize here that although our motion discriminator acts on sequences, the hand pose model is a per-frame model.

\noindent\textbf{Properties of 2.5D.} We briefly touch on some useful properties of the chosen representation which we will reference later. First, both the predicted values $\V{J}^{2D}, \V{z}^{r}$ are inherently \emph{bounded}. The 2D keypoints must lie within the image, hence are constrained by the image size. The root-relative depth is naturally bounded by the skeletal structure of the hand, where its maximum absolute value is the maximum distance between the root joint and the corresponding keypoint. However, a full 3D representation is unbounded as the 3D joint skeleton can lie anywhere within the camera coordinate frame. The second useful property is \emph{translation invariance}. No matter where in world coordinates the 3D pose lies, the camera intrinsics are adjusted such that they project onto the same image plane, resulting in the same 2D joint positions. Similarly, translating the 3D pose does not affect $\V{z}^r$, as the translation is applied to all keypoints.

\subsection{Adversarial motion model}
The goal of the adversarial learning is to make use of additional unlabeled video data on which $M$ can be further trained on. To this end, we introduce a discriminator $D: \R^{Sx21x3} \longrightarrow \R $ which takes in a sequence of hand joints $\V{J} \in \R^{S x 21 x 3}$ and outputs a value in $[0,1]$, classifying between real and predicted sequences. Formally, $D$ is trained to optimize the following objective function:
\begin{equation}
\begin{split}
\label{eq:d_objective}
    \loss_D = \E_{\V{J}\sim p_R} [(1 - D(\V{J}^{3D}))^2] + \E_{\hat{\V{J}}\sim p_M} [D(\hat{\V{J}}^{3D})^2]
\end{split}
\end{equation}
Here, $p_R$ is the distribution of real hand motion, whereas $p_M$ is the distribution of the prediction of our hand pose model. We use the LSGAN \cite{mao2017least} loss function due to its stability, following \cite{kocabas2019vibe}. To train $M$ on unlabeled videos, we update it to confuse the discriminator:
\begin{equation}
    \begin{split}
    \label{eq:adv_loss}
        \loss_{MM} = \E_{\V{I}^{1:K}\sim p_I} [(1 - D(M(\V{I}^{1:K})))^2]
    \end{split}
\end{equation}
Where $\V{I}^{1:K}$ are the $K$ frames of a given unlabeled image sequence. We 
define $M(\V{I}^{1:K}) = [M(\V{I}^1), \dots M(\V{I}^K)]$, i.e $M$ returns a per-frame prediction.

\noindent\textbf{Final loss.} $M$ is jointly trained on supervised data via $\loss_{J^{2D}},\loss_{z^r}$ and on unlabeled data using $\loss_{MM}$. The final loss is 
\begin{equation}
    \loss = \loss_{J^{2D}} + \lambda_z \loss_{z^r} + \lambda_{MM} \loss_{MM}
\end{equation}
Where $\lambda_i$ is a weighing factor.

\subsection{Design choices}
\label{sec:design}
In Sec.~\ref{sec:intro} we have alluded to the difficulties in translating motion models from the full body literature to the hand pose setting. Since we believe that the proposed adversarial setting is a fruitful step into this direction and can lead to further improvements in future work, we briefly discuss our design choices for the adversarial learning step and the underlying model. Each of them is backed up by empirical evidence which is provided in Sec.~\ref{sec:ablation}. For the sake of clarity, we also discuss methods that have worked in related fields but lead to detrimental results in our experiments. 

\noindent\textbf{Spectral normalization \cite{miyato2018spectral}.} Applying spectral normalization to the weights of the discriminator has been shown to stabilize training and generate images of higher quality when comparing to previous training stabilization techniques. During the course of our experiments, we have found that spectral normalization does not yield clear improvements. Therefore omit it from our discriminator.

\noindent\textbf{Network depth.} We represent a sequence of predictions as a matrix. This means that adjacent values do not necessarily share a spatial relationship. Prior work alleviate this by introducing special spatial encoding layers \cite{hernandez2019human} or increasing the receptive field of the network to be large enough \cite{kaufmann2020convolutional}. We determined that two residual blocks and an encoding and decoding layer performed best.

\noindent\textbf{Batch normalization \cite{ioffe2015batch}.} Batch normalization speeds up training of neural networks. During training, it normalizes its input feature maps to have a zero mean and unit variance distribution across the batch followed by learnable scaling and shifting. 
At test time, it makes use of the aggregate batch statistics which was acquired during training for normalization. We have made the observation that $M$ would diverge during adversarial learning when the current batch statistics are used. 
Therefore we use the aggregate batch statistics after the pre-training phase. %

\noindent\textbf{Sequence length.} We experiment with various sequence length. We test 1, 16, 32 and 64 frames which correspond roughly to 1/30, 0.5, 1, 2 seconds respectively. Longer sequence lengths allows the discriminator to decide based on more information. However, it increases computation time and dimensionality, which may lead to overfitting. In our experiments using 16 frames yielded the best performance.

\noindent\textbf{Joint representation.} $M$ predicts the 2.5D representation of keypoints which can be easily converted to 3D (Eq. \ref{eq:25d3d}). It is unclear however, which representation is best used for the motion modeling task. Whereas the 3D representation directly encodes the task at hand, the 2.5D representation is more constrained. Our analysis shows that $M$ more effectively learns when the discriminator uses the same representation, hence we choose 2.5D over 3D.

\noindent\textbf{Data augmentation.} Augmenting the input to the discriminator has been shown to improve performance \cite{karras2020training}. As the input are joint skeleton sequences, we perform geometric augmentation. The 2.5D representation is translation invariant therefore we resort to rotations. Care needs to be taken to apply the augmentation consistently across the sequence as they should not incorrectly change the statistics of a given sequence. Hence, given a sequence of poses, we rotate it around the root of the first joint skeleton of the sequence. This ensures that the relative change in motion between two frames of a sequence remains the same. Because out-of-image-plane rotations could result in joint skeletons being behind the camera, we only perform rotations around the z-axis of the camera.
\section{Implementation}
Our pose model uses a ResNet-18 \cite{he2016deep} backbone that takes a 256 $\times$ 256 RGB image and outputs the 2.5D representation of keypoints \cite{iqbal2018hand}. Following \cite{spurr2020weakly}, we implement a refinement network to stabilize the predictions of the absolute depth. We pre-train the hand pose model on the labeled data twice, once using the current batch statistics and once using the aggregate. We then jointly train on both labeled data via $\loss_{J^{2D}}, \loss_{z^r}$ and unlabeled data via $\loss_M$. Our discriminator architecture is implemented as a CNN with residual connections that takes a sequence of joint predictions and outputs a valid/invalid motion flag. It is trained on the hand motion data of the respective dataset. Exact training and architecture details can be found in the appendix.
\section{Experiments}
\begin{figure*}[t]
    \centering
    \includegraphics[width=0.9\textwidth]{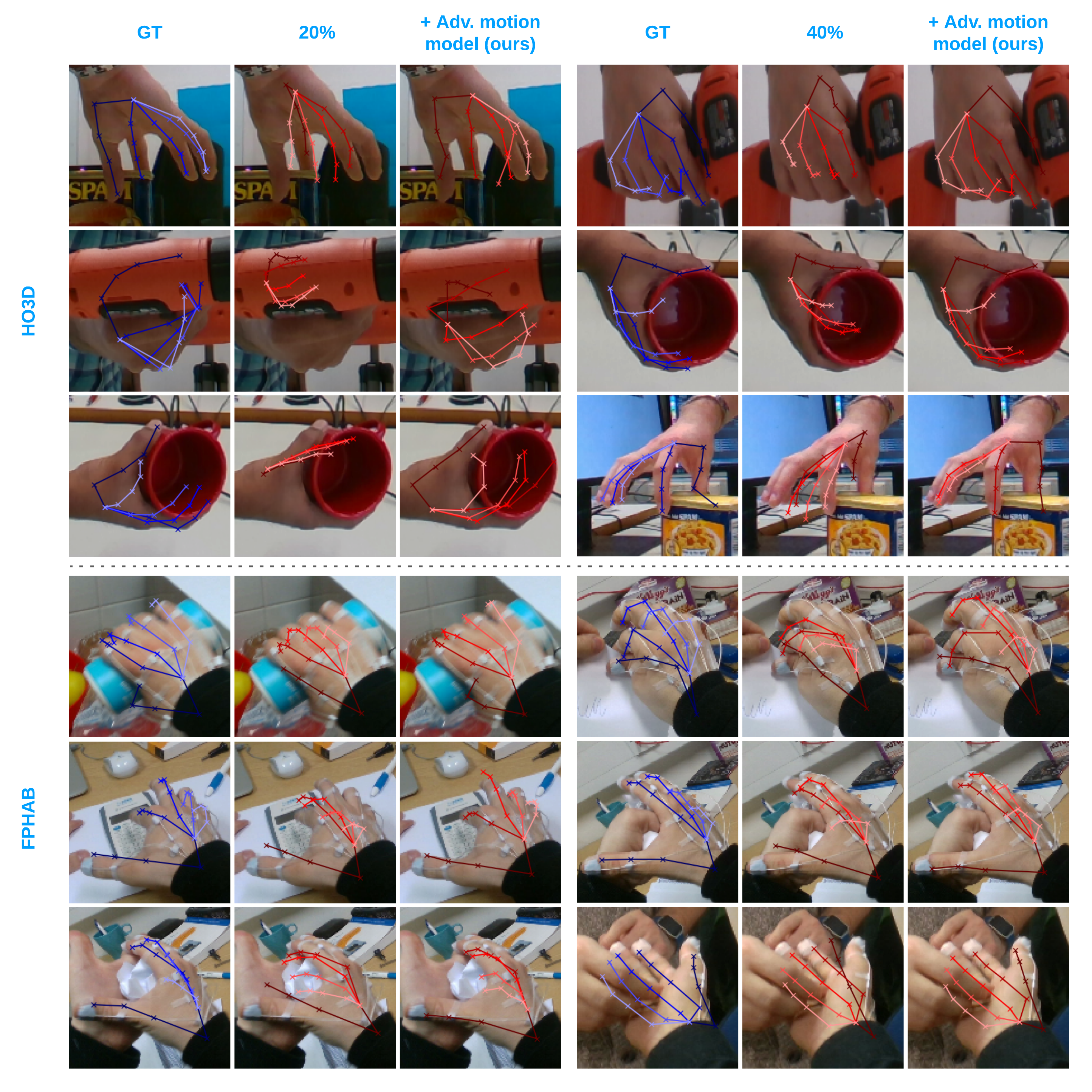}
    \vspace{-3mm}
    \caption{\textbf{Qualitative results.} (top row) HO3D (bottom row) FPHAB. For each $3\times3$ block, the left column shows the ground-truth, the middle shows the prediction result of the baseline model train on 20\% (left blocks) or 40\% (right blocks) labels, the right column shows predictions of the model that learned from on additional unlabeled videos using our proposed adversarial motion model. The predictions are more accurate, biomechanically plausible and fit the displayed hand better.} %
    \label{fig:qual_results_all}
\end{figure*}
We analyze how the proposed motion model leads to improved pose prediction using unlabeled video data.

\subsection{Datasets}
To quantitatively evaluate our framework, we require access to fully labeled \emph{sequential} data of 3D hand poses. Currently, only two datasets that fulfill this requirement.

\noindent\textbf{FPHAB \cite{garcia2018first}} contains egocentric sequences of RGB-D videos. The motion performed capture a wide range of interactions, such as hand-object as well as hand-hand. It contains  ground-truth annotations of the hand pose, acquired with magnetic sensors strapped to the hand. Following \cite{hasson2020leveraging}, we report our results on the \textit{action} split.

\noindent\textbf{HO3D \cite{hampali2020honnotate}} contains 3rd person view sequences of hand-object interactions. It was captured using a multi-RGB-D-camera set up. An earlier version of this dataset exists \cite{hampali2019ho} which was used by \cite{hasson2020leveraging}. Here, we use the newest version.

\subsection{Evaluation metrics}
\label{sec:eval_metrics}
\noindent\textbf{FPHAB.} We report the mean joint error in mm for both the absolute and root-relative pose. The former is the quantity of interest, however it heavily depends on correctly estimating the distance to the camera. As this is a severely ill-posed for monocular RGB, one may encounter large absolute pose errors despite predicting correct articulations. Hence we also report the root-relative error to quantify the articulation error in absence of the camera distance estimate.

\noindent\textbf{HO3D.} An online submission system returns the absolute, scale/translation (ST) aligned as well as the procrustes-aligned EPE in cm. To be consistent, we report the former two and convert them to mm. %

\subsection{Settings}
We explore two different semi-supervised protocols. To the best of our knowledge, the only related work that explores a similar setting to ours is \cite{hasson2020leveraging}. Therefore all of our semi-supervised experiments compare to their method. To compare directly, \textbf{Protocol 1} follows \cite{hasson2020leveraging} labeling precisely, where the labels of each sequence are sampled uniformly, starting from the first frame. However, as frames within a sequence tend to be similar, we cannot extrapolate the performance of our method on completely unlabeled videos. Therefore we introduce \textbf{Protocol 2} where we sample \textit{entire} labeled videos, keeping the remaining videos unlabeled. This setting is more challenging, as there exists a much bigger difference between the labeled and unlabeled data and as such more closely simulates truly unseen videos. More details for both protocols are found in the appendix.

\subsection{Ablation study}
\label{sec:ablation}

\begin{table}[th]
\small
\centering
\begin{tabular}{lcc}
\toprule
\multirow{3}{*}{Ablation study} & \multicolumn{2}{c}{3D hand pose estimation} \\ 
               & \multicolumn{2}{c}{EPE (mm)}           \\ 
               & Absolute $\downarrow$          & Root-relative $\downarrow$    \\ 
\midrule
\multicolumn{3}{c}{Motion model type}                   \\
\midrule
Temporal prior      & 27.44         &   13.31           \\
Data-driven         & \bf 23.86     &   \bf 11.20       \\
\midrule     
\multicolumn{3}{c}{Spectral normalization}              \\ 
\midrule
With           & 24.42              & \bf 10.93             \\ 
Without        & \bf 23.52          &  11.02         \\ 
\midrule
\multicolumn{3}{c}{Network depth}                       \\ 
\midrule
1              & 23.86              & 11.20             \\ 
2              & \bf 23.52          & \bf 11.02         \\ 
4              & 25.57              & 12.35             \\ 
8              & 26.94              & 12.89             \\ 
\midrule
\multicolumn{3}{c}{Sequence length}                     \\ 
\midrule
1              & 27.16              & 12.80             \\
16             & \bf 23.52          & \bf 11.02         \\ 
32             & 25.03              & 11.52             \\ 
64             & 25.48              & 11.58             \\ 
\midrule
\multicolumn{3}{c}{Data augmentation}                   \\ 
\midrule
With           & \bf 23.52          & \bf 11.02         \\ 
Without        & 24.46              & 11.30             \\ 
\midrule
\multicolumn{3}{c}{Keypoint representation}             \\ 
\midrule
3D             & 26.62              & 13.59             \\ 
2.5D           & \bf 23.52          & \bf 11.02         \\ 
 
\bottomrule
\end{tabular}
\caption{Full ablative evaluation motivating our design choices. We evaluate on FPHAB in the $20\%$ labeled setting of Protocol 2.}
\label{tbl:ablation}
\end{table}

We empirically present our results that motivated the design choices in Sec.~\ref{sec:design}. The complete evaluation for \emph{all} ablative experiments is presented in \tabref{tbl:ablation}. These are done using Protocol 2 and $20\%$ labeled data on FPHAB.

\noindent\textbf{Motivating data-driven motion models.}
Prior work \cite{hmrKanazawa17} make use of adversarial methods to improve single frame predictions which could be used to leverage unlabeled \emph{images}, yielding a pose prior. Alternatively, simple temporal priors could be employed to take advantage of unlabeled \emph{videos}. We explore both viable alternatives to the proposed data-driven motion model. For the pose prior, we change the length of the sequence that we feed to the discriminator to 1. This yielded $27.16$mm MPJPE. Making use of a temporal smoother that penalizes non-smooth motion ($||x_t - x_{t-1}||_2$) performed worse, achieving $27.44$ mm MPJPE. However, both approaches perform worse than the proposed motion model which yields $23.52$mm EPE.

\noindent\textbf{Design choices.} 
\tabref{tbl:ablation} shows the empirical evidence motivating the design choices of the motion model. Spectral normalization and deeper network models lead to an increase in training time and error whereas using 16 frames lead to optimal results. Our proposed data augmentation scheme produced a decrease of $0.94$ mm. Lastly, making use of the 2.5D keypoint representations yields a $3.1$ mm drop in error. 

The analysis of design of the adversarial motion model training framework yielded a total reduction in absolute error from $28.47$ mm to $23.52$ mm, a decrease of $4.95$ mm. %

\subsection{Semi-supervised adversarial learning}
\begin{table}[b]
\small
\centering
\begin{tabular}{lc}
\toprule
 FPHAB      & Absolute MPJPE $\downarrow$ (mm)      \\
\midrule
Tekin et al. CVPR'19 \cite{tekin2019ho} & 15.8 \\
Hasson et al. CVPR'20 \cite{hasson2020leveraging} & 15.7          \\
Ours          & \bf 15.5        \\
\midrule
HO-3D      & ST-aligned MPJPE $\downarrow$ (mm)      \\
\midrule
Hasson et al. CVPR'19 \cite{hasson2019learning} & 31.8          \\
Hampali et al. CVPR'20 \cite{tekin2019ho} & 30.4 \\
Ours          & \bf 24.5        \\
\bottomrule
\end{tabular}
\caption{Prior work comparison in the fully supervised setting. Please note that \cite{hasson2019learning} and \cite{hasson2020leveraging} are differing works.}
\label{tbl:fully_sup}
\end{table}

\begin{figure*}
     \centering
    \begin{subfigure}[b]{0.32\textwidth}
         \centering
         \includegraphics[width=1\columnwidth]{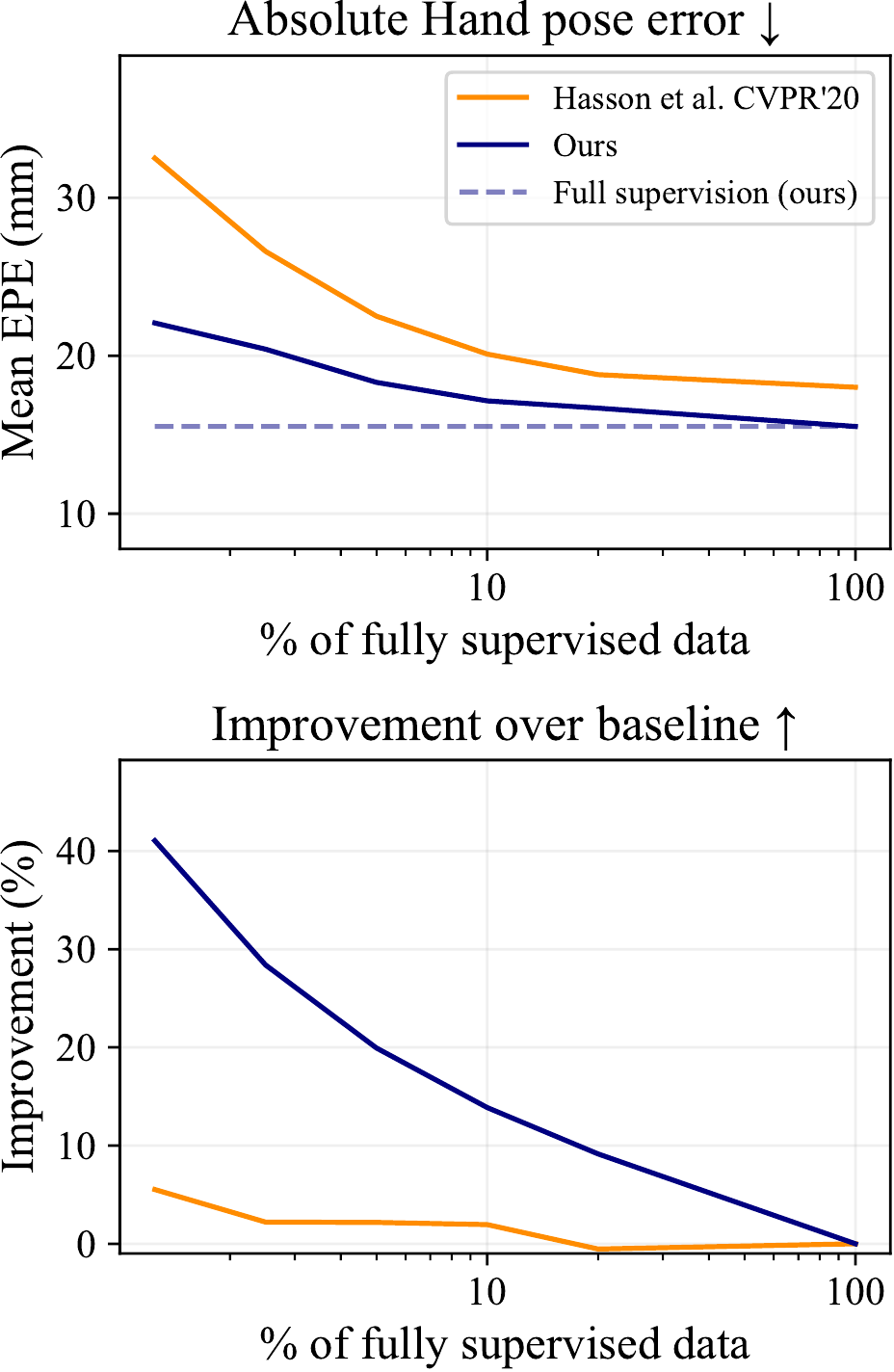}
         \caption{FPHABA - Protocol 1}
         \label{fig:protocol1}
     \end{subfigure}%
     \hspace{1mm}%
     \begin{subfigure}[b]{0.32\textwidth}
        \centering
        \includegraphics[width=1\columnwidth]{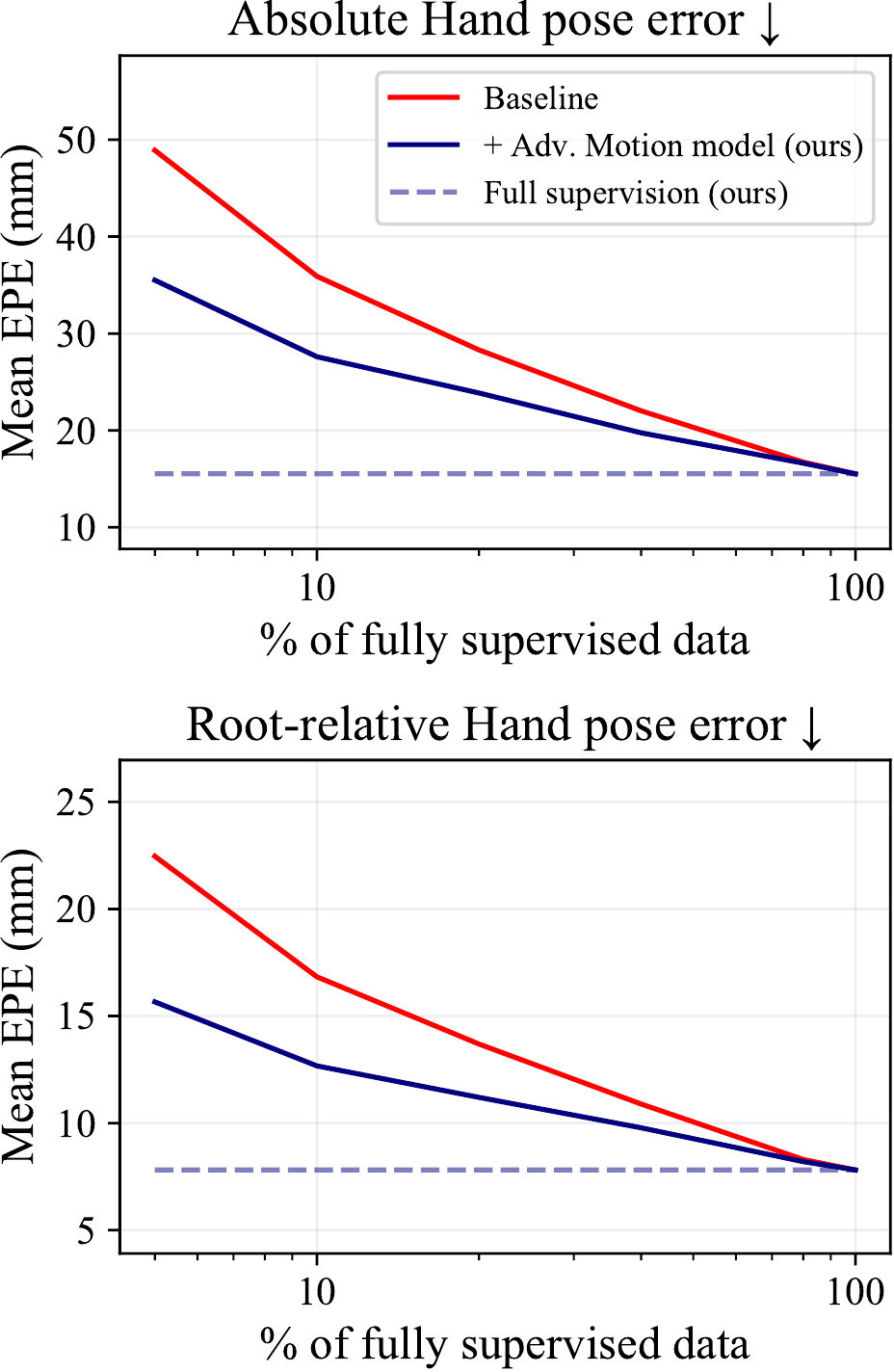}
        \caption{FPHABA - Protocol 2}
        \label{fig:protocol2_fphab}
     \end{subfigure}%
     \hspace{1mm}%
     \begin{subfigure}[b]{0.32\textwidth}
         \centering
         \includegraphics[width=1\columnwidth]{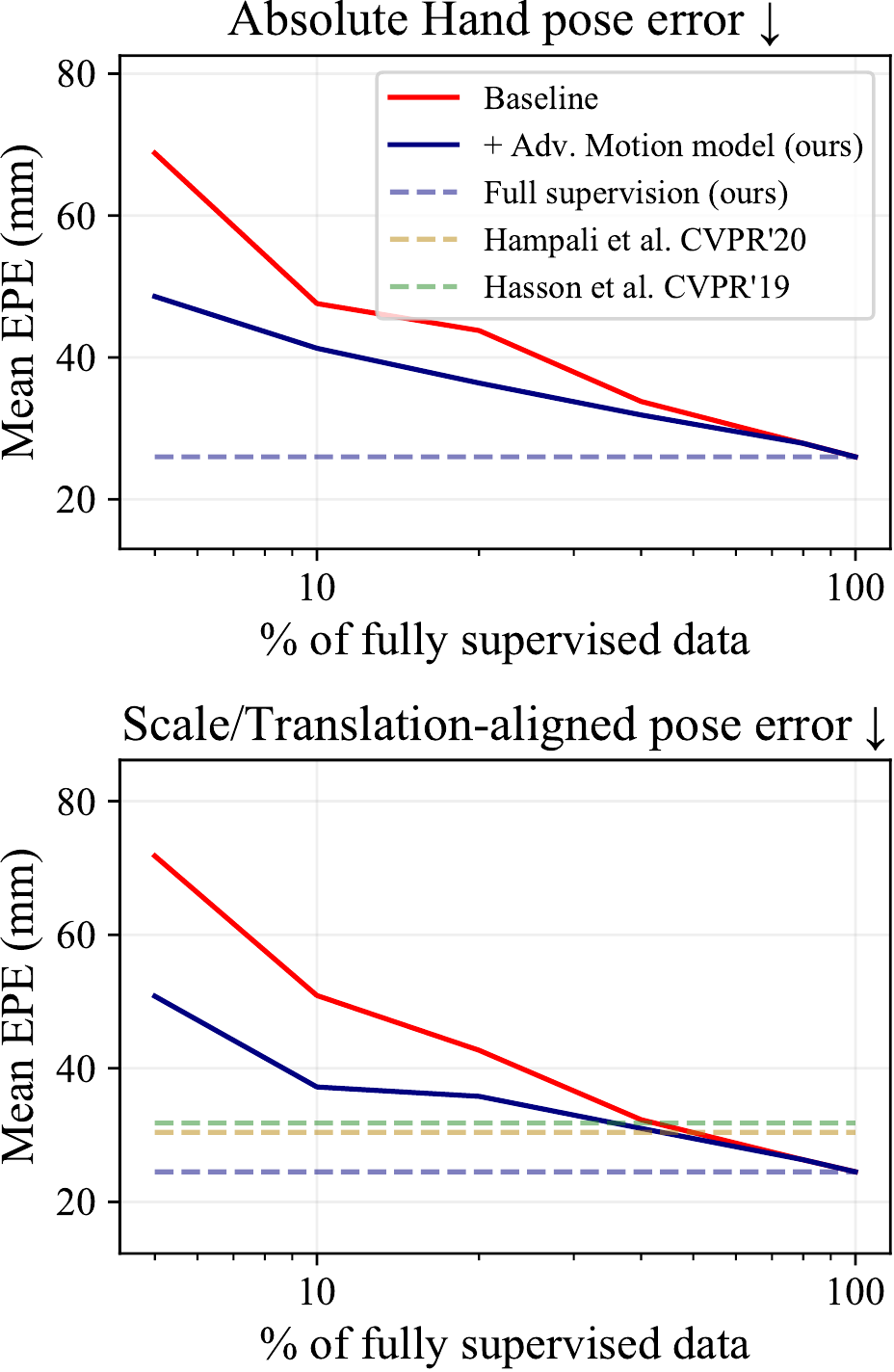}
         \caption{HO-3D - Protocol 2}
         \label{fig:protocol2_ho3d}
     \end{subfigure}
     \vspace{-1mm}
    \caption{\textbf{Comparison with related work.} \figref{fig:protocol1} Our method outperforms \cite{hasson2020leveraging} for all labeling percentages. In addition our method yields greater improvement over the respective baseline for all labeling percentages. \figref{fig:protocol2_fphab} leveraging unlabeled \emph{sequences} is more challenging, which can be seen by the overall higher error rate. We observe improvements for both absolute, as well as root-relative hand pose error. \figref{fig:protocol2_ho3d} demonstrates a similar trend on HO3D.}
    \label{fig:protocol2}
\end{figure*}

Next we evaluate the proposed approach under different amount of labeled sequences and under both protocols.

\noindent\textbf{Fully-supervised performance.} We evaluate the performance of our baseline model in the fully supervised setting on FPHAB and HO3D. The results are presented in \tabref{tbl:fully_sup}. For FPHAB, we compare with \cite{tekin2019ho, hasson2020leveraging}, demonstrating that our baseline  performs on par with prior work. For \cite{hasson2020leveraging}, we report their fully supervised "Hands only" result, as we focus on hand pose too. For HO3D, we compare to \cite{hampali2020honnotate, hasson2019learning}. For \cite{hasson2019learning}, the results are obtained from \cite{ho3dwebsite}. %

\noindent\textbf{Protocol 1.} We report our results on in \figref{fig:protocol1}, comparing to \cite{hasson2020leveraging}. As they make use of an early version of HO3D, we compare only on FPHAB. We observe that we outperform their approach for all percentage of supervision used. In addition, the improvement of our method over our baseline is larger than \cite{hasson2020leveraging} over theirs. We note here that the semi-supervised results of \cite{hasson2020leveraging} require access to the ground-truth mesh of the object, hence their reported results are for the model that predict hand and object pose. Designing a semi-supervised ``hands only'' variant is non-trivial due to hard-coded reliance on the to the ground-truth mesh to compute the photometric consistency loss.\footnote{Confirmed by authors via private correspondence.}
Increasing the amount of labeled information available, both approaches converge to their fully supervised results as is expected (cf.~\tabref{tbl:fully_sup}). We note here that both methods complement each other and can be combined. Whereas \cite{hasson2020leveraging} reasons on a pixel-level, we reason on a motion level. Future work could explore a combination of both approaches.

\noindent\textbf{Protocol 2.} We present our quantitative results on FPHAB in Fig.~\ref{fig:protocol2_fphab} for both the absolute and root-relative pose. We compare a baseline method, which does not make use of the adversarial motion model, to our approach. Compared to protocol 1 the models exhibit larger discrepancies in error between the lower labeling percentages and the fully supervised case. This is indicative of the increased difficulty of Protocol 2, as the unlabeled data contains completely unseen imagery. We observe that when labeled data is scarce, our proposed adversarial loss has a much larger impact on the final performance. This is evident when comparing the performance of our method in the lower label regime with the baseline model of higher percentages. For example, our method with $10\%$ labeled data achieves $27.60$ mm error, outperforming the baseline with $20\%$ labeled data at $28.30$ mm error (cf. \figref{fig:protocol2_fphab}). This suggests that the proposed motion model is effective in leveraging unlabeled videos. Fig.~\ref{fig:protocol2_ho3d} shows our results on HO3D given by the online evaluation system. We observe similar trend as with FPHAB.

\section{Conclusion}
In this work we have proposed to leverage an adversarial motion model to improve the task of hand pose estimation to learn from unlabeled data. Due to lack of related work, we provided an overview of the essential components to make use of a simple but effective motion model in order to leverage unlabeled video for hand pose estimation learning. The goal is to provide a foundation on which future work combining both methodologies can build upon. We evaluated our proposed method in two challenging settings and demonstrated that significant improvements can be gained. Future work will look into extending the proposed methodology across datasets, making use of in-the-wild videos such as YouTube3DHands \cite{kulon2020weakly}.

{\small
\bibliographystyle{ieee_fullname}
\bibliography{egbib}
}
\end{document}